\documentclass[letterpaper, 10 pt, conference]{ieeeconf}  
\usepackage[utf8]{inputenc}
\usepackage{amsmath, amsfonts}
\usepackage{graphicx}
\usepackage{xcolor}
\usepackage{wrapfig}
\usepackage{adjustbox}
\usepackage{xspace}
\usepackage{epstopdf}
\usepackage{xcolor}
\usepackage{float}
\usepackage{soul}
\usepackage{hyperref}
\usepackage{siunitx}
\usepackage{algorithm}
\usepackage{algcompatible}
\usepackage[capitalize]{cleveref}
\usepackage{booktabs}
\usepackage{multirow}
\usepackage[para]{threeparttable} 
\usepackage{array}
\usepackage{colortbl}
\usepackage{siunitx}
\usepackage{pifont}

\usepackage{eqexpl}
\eqexplSetDelim{=}

\usepackage{algpseudocode}
\usepackage{array}
\usepackage{tabularx}

\usepackage{wrapfig}
\usepackage[small]{caption}
\usepackage{subcaption}
\usepackage[noadjust]{cite}
\usepackage{comment}

\IEEEoverridecommandlockouts 

\newcommand{\systemname}{\textit{}\xspace}

\title{\systemname Design, Mapping, and Contact Anticipation with 3D-printed\\ Whole-Body Tactile and Proximity Sensors}
\author{Carson Kohlbrenner, Anna Soukhovei, Caleb Escobedo,\\ Nataliya Nechyporenko, and Alessandro Roncone
\thanks{All authors are with the Human Interaction and RObotics [HIRO] Group, University of Colorado Boulder, 1111 Engineering Drive, Boulder, CO USA. This work is partially supported by NSF FW-
HTF-R grant \#2222952.
{\tt\small name.surname@colorado.edu}.}
}

\begin{document}
\maketitle

\begin{abstract}
Robots operating in dynamic and shared environments benefit from anticipating contact before it occurs. We present GenTact-Prox, a fully 3D-printed artificial skin that integrates tactile and proximity sensing for contact detection and anticipation. The artificial skin platform is modular in design, procedurally generated to fit any robot morphology, and can cover the whole body of a robot. The skin achieved detection ranges of up to 18 cm during evaluation. To characterize how robots perceive nearby space through this skin, we introduce a data-driven framework for mapping the Perisensory Space---the body-centric volume of space around the robot where sensors provide actionable information for contact anticipation. We demonstrate this approach on a Franka Research 3 robot equipped with five GenTact-Prox units, enabling online object-aware operation and contact prediction.
\end{abstract}


\section{Introduction}

The human skin, the body’s largest organ, contains dense networks of afferent nerves that convey contact information to the brain \cite{corniani2020tactile, ackerley2014touch}. In addition to direct touch, humans rely on multisensory integration---including visual and auditory cues---to construct a representation of the nearby interactable space \cite{roncone2015learning,roncone2016peripersonal}. This representation, known as PeriPersonal Space (PPS), enables the anticipation of contact before physical interaction occurs \cite{bertoni2025computational,serino2022peripersonal, cooke2003complex}. The PPS is fundamental for moving in contact-rich environments because it provides a predictive buffer that allows humans to avoid harmful collisions and deliberately engage in contact. For robots expected to similarly operate in contact-rich environments, establishing an artificial equivalent of PPS is a key step toward achieving anticipatory spatial awareness.

Whole-body artificial skin can enable robots to perceive contact, but few works use whole-body artificial skin to directly measure the \textsl{surrounding space} of a robot from a body-centric reference frame. Additionally, design challenges such as scalability, reproducibility, calibration, and lack of a holistic understanding of how to utilize the surrounding space with whole-body artificial skins have made implementing the PPS difficult in practice. These challenges call for a renewed focus on the assumptions and hardware used for establishing the PPS. To this end, we propose here the term \textbf{PeriSensory Space (PSS)}, defined as the volume of space around the body that can be actively sensed, as a characterizing metric for the PPS. Unlike the PPS, the PSS is determined by the observable space of its onboard sensors and is not dependent on the robot's action. We propose that the PSS serves as the information gateway for establishing and updating the PPS online. Artificial skin solutions that sense proximity \cite{rupavatharam2023ambisense,fan2021aurasense, escobedo2021contact,kim2024armor} or hybrid touch and proximity sensing \cite{giovinazzo2024cyskin,zhou2023tacsuit,cheng2019comprehensive, yang2024digital} can observe the nearby space of a robot and establish a PSS. 

To overcome the hardware challenges of large-scale sensor development, a pipeline called \textit{GenTact} was introduced as a low-cost and open source way to design and fabricate whole-body skins \cite{kohlbrenner2024gentact}. Artificial skins made with this pipeline are procedurally generated to fit a specific robot and are fully three-dimensionally (3D) printed, requiring low technical expertise and material cost to reproduce at scale \cite{kohlbrenner2024gentact}. Although promising, \textit{GenTact}-made skin sensors have only been demonstrated for binary contact sensing, and whole-body 3D-printed artificial skins that can detect both touch and proximity remain unexplored.

\begin{figure}
    \centering
    \includegraphics[width=\linewidth]{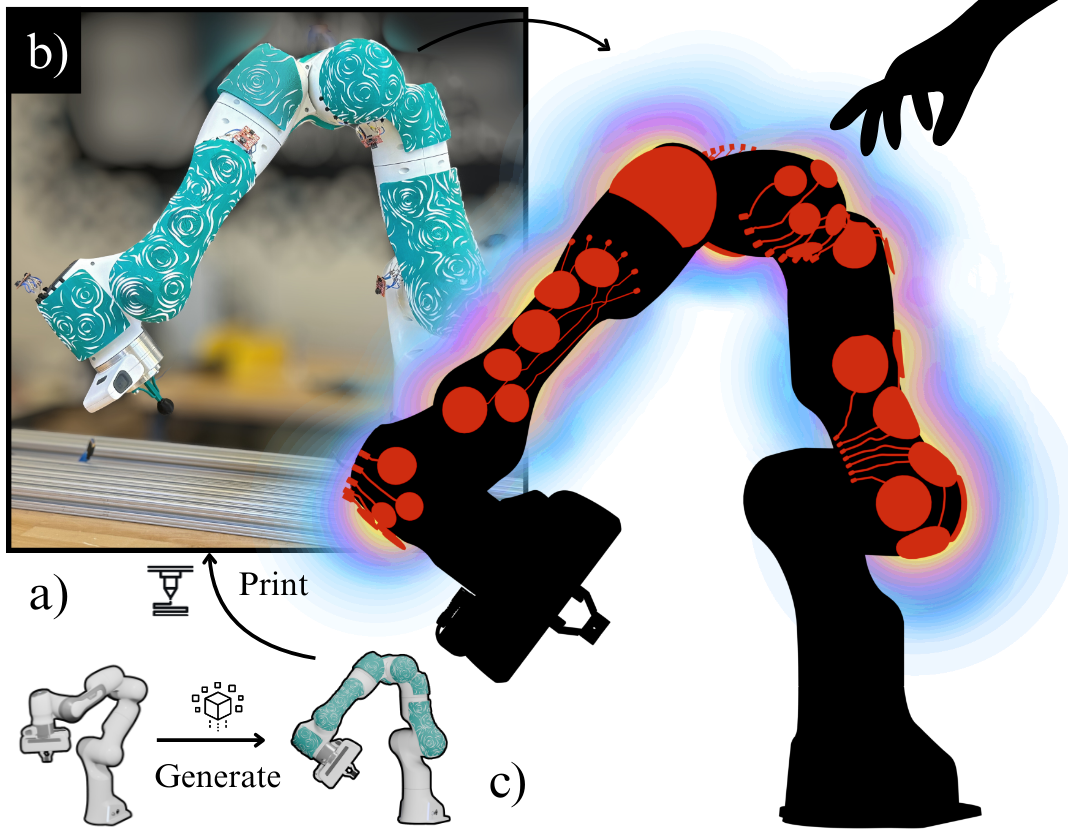}
    \caption{a) \textit{GenTact-Prox} whole-body artificial skins are digitally generated to fit a robot's morphology and 3D-printed with functional tactile and proximity sensors. b) Five unique skin units were deployed on the FR3 robot for evaluation. c) An ``x-ray" view of the individual sensors nested within the \textit{GenTact-Prox} skin units surrounded by the sensor's Perisensory Space (PSS).}
    \label{fig:cover}
    \vspace{-1.5mm}
\end{figure}

To make the artificial PSS widely accessible through the advantages of 3D printing, we introduce \textit{GenTact-Prox}, an extension of the \textit{GenTact} pipeline for making whole-body artificial skins that are sensitive to both tactile and proximity stimuli (\cref{fig:cover}). 
Inspired by recent work in 3D printed capacitive sensors, we procedurally define the geometry of capacitive conductive plates to be embedded within the outer layer of the skin. By embedding the sensors within the skin, the plates can be sensitively tuned to detect the weaker proximity effect of nearby objects without saturating from direct contact. Each artificial skin is designed to conform to a target robot's shape, ensuring accurate sensor placement relative to underlying links.
Our artificial skin solution allows the capacitive sensors to be shaped and distributed both uniformly or non-uniformly to optimize for sensing coverage and design flexibility. However, the non-uniform distribution and shape of our capacitive proximity sensors contradict with common assumptions made by the literature to overcome the sensor's non-linear behavior \cite{Ye2020ARO}.

Prior capacitive proximity sensing methods often rely on well-defined electrode structures to approximate a sensed object's distance \cite{Ye2020ARO}, an approach that does not extend to arbitrary sensor layouts. A data-driven strategy using an ensemble of machine learning models is proposed to directly capture the signal–space relationship surrounding a given robot and skin configuration. We use a robot to hover a known object over each of the sensors to teach the ensemble how to localize nearby objects, and then use the prediction error between models in the ensemble to quantify a learned PSS that is specific to a robot with arbitrary sensor distributions. We show that the prediction error of the ensemble is correlated to the space observable by the sensors, bypassing the need for assumptions on the sensor's shape and distribution.

In summary, our contributions are (i) \textit{GenTact-Prox}, a 3D printed whole-body artificial skin solution that senses tactile and proximity stimuli, and (ii) a machine learning method for mapping the Perisensory Space (PSS) of a robot covered by arbitrarily distributed onboard capacitive sensors.
We demonstrate our approach by characterizing and deploying five \textit{GenTact-Prox} skin units on a Franka Research 3 (FR3) arm, mapping its respective PSS, and performing collision-compliant control. 
We envision that this work will lower the barrier to entry for making interactive robots that can anticipate contact in the space around them. The presented procedural operations and printable STL files of the sensors are open-sourced and available at \url{https://hiro-group.ronc.one/gentacttoolbox}

\section{Related Work}

\subsection{3D Printed Capacitive Sensors}
Conductive filaments have been used in recent works to print functional electronics within 3D printed objects \cite{he2021modelec,burstyn2015printput, savage2014series}. 3D printed electronics are not constrained to flat surfaces, and tools that automatically generate internal circuits in generic 3D objects have been explored \cite{savage2014series, palma2024capacitive, bae2023computational}. Capacitive sensors can detect direct contact with exposed electrodes or estimate proximity and are commonly used in 3D printed electronics due to their low cost and relative simplicity in design \cite{schmitz2015capricate, schmitz2019trilaterate}. However, 3D printed capacitive sensors have only been limited to small-scale examples on robots \cite{xiao2025fully, taylor2024fully}. In this work, we integrate the tactile and proximity sensing capabilities of 3D printed capacitive sensors into the whole body.

\subsection{Modeling the Peripersonal Space in Robots}
The PPS is a term derived from neuroscience research that describes the interactable space surrounding the body \cite{bertoni2025computational}. The implications of the PPS have motivated roboticists to implement artificial analogs in robots, which have been used for tasks such as collaborative games \cite{rozlivek2024harmonious}, collision avoidance \cite{escobedo2021contact}, and teleoperation \cite{poignant2024teleoperationroboticmanipulatorperipersonal}. Such analogs can only be informed by the observations made in the receptive field of onboard sensors \cite{roncone2016peripersonal}. Structured sensors with known receptive fields can be mapped to the body-centric PSS through calibration and forward kinematics \cite{roncone2016peripersonal, watanabe2021self}. However, mapping the PSS with heterogeneously structured capacitive sensors becomes convoluted due to their unknown receptive fields. Our approach leverages machine learning to overcome this lack of structure and map the PSS of heterogeneous capacitive sensors distributed over the whole body of a robot.

\section{Methods}
\label{sec:methods}
The following section covers the design, fabrication, and implementation of a \textit{GenTact-Prox} skin unit. Each step presented is repeated for each modular unit. 

\subsection{Procedural Generation} 

Each skin unit is procedurally generated using Blender geometry nodes to promote rapid design iteration, reduce the modeling expertise required to modify the units, and support design flexibility for sensor distribution. Procedural generation is a series of $m$ operations $P_{1:m}$ that takes in a geometric set of vertices, edges, and faces, $G_0=(V_0,E_0,F_0)$, as well as a set of $n$ user defined parameters $A=\{a_1, a_2, ...,a_n\}$ to create a new geometry $G$. Once the procedural operations are defined, a designer only needs to provide the robot model and tune the design parameters to generate a functional skin unit. Additionally, vertex weights represented as a heat map $H$ can be passed into the operations to specify specific regions of the geometry to execute on. Our algorithm is defined to create a skin unit model $G_{\text{SU}}$ from a robot 3D model $G_{0}$ as follows:
\begin{equation}
    P(G_{0}, A, H_{\text{skin}}) = G_{\text{SU}}
    \label{eq:procedural}
\end{equation}

\begin{figure}
\vspace{2mm}
    \centering
    \includegraphics[width=\linewidth]{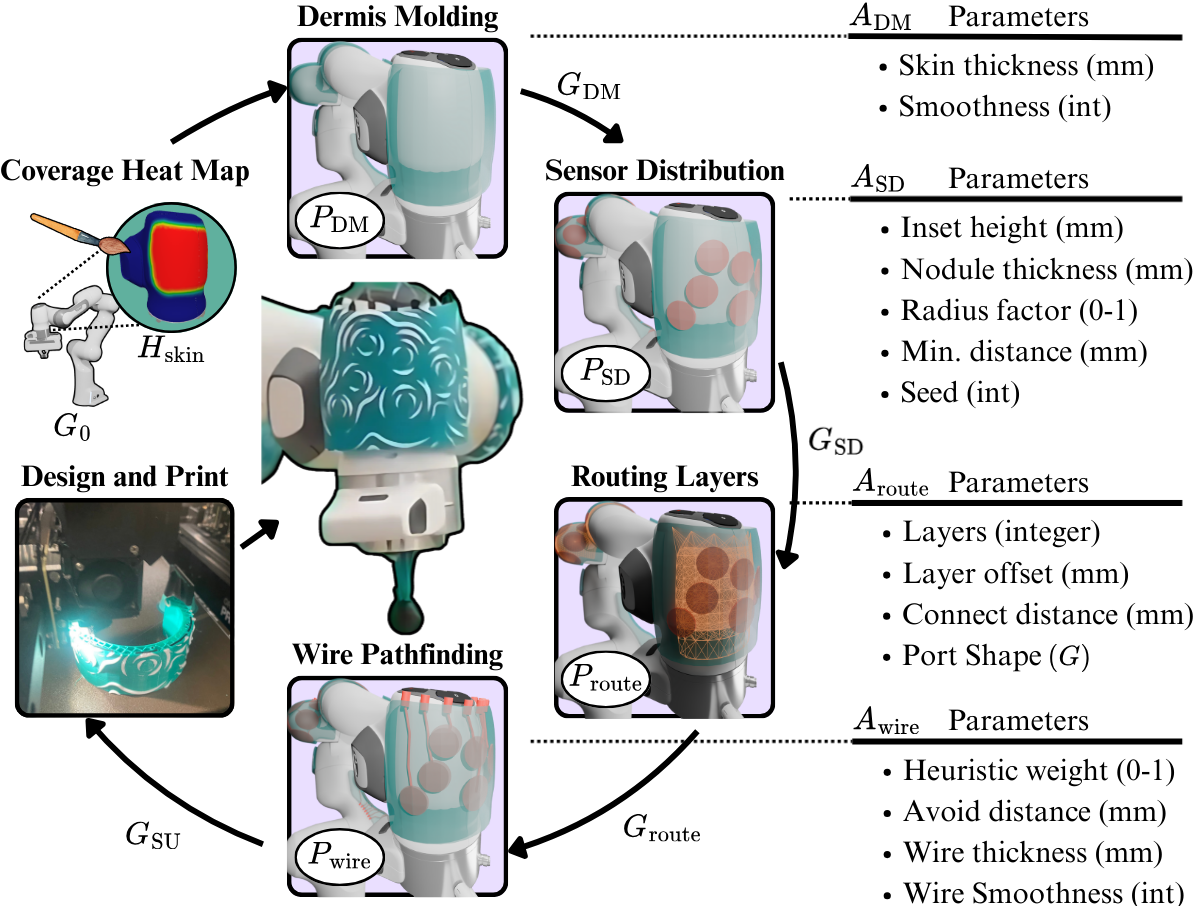}
    \caption{The full procedural design stage can be categorized into four main processes: $P_{\text{DM}}\rightarrow P_{\text{SD}} \rightarrow P_{\text{route}} \rightarrow P_{\text{wire}}$. The series of operations molds the dermis layer, distributes the sensors within the dermis, routes a graph of the possible space that wires should occupy, and finally solves for non-overlapping wires between each connection port and sensor. Each process depends on its respective parameters, and the swirl design on the skin serves as a visual reference for the embedded sensors.}
    \label{fig:design-chart}
    \vspace{-1.5mm}
\end{figure}

The process $P$ consists of four operations: (1) a dermis molding operation $P_{\text{DM}}$ that generates the structural layer to house the electronics and provide collision compliance; (2) a sensor distribution operation $P_{\text{SD}}$ that places disk nodules within the dermis; (3) a routing operation $P_{\text{route}}$ that creates connection ports and maps possible wire paths; and (4) a wiring operation $P_{\text{wire}}$ that connects the ports to their respective nodules without overlap. A step-by-step chart of each operation is shown in \cref{fig:design-chart}. The operations are formally defined as follows:

$P_{\text{DM}}$ molds the dermis layer by discarding unweighted vertices (i.e. $V_{\text{skin}}=\{v_i\in V \mid h_i =1\}$ for $h_i\in H_{\text{skin}}$), extrudes all remaining valid faces outward along their local normal vectors to a set thickness, and finally smooths the outer edges to counter any broken geometry found in the original model. The dermis layer is smoothed by fitting a Catmull-Rom spline along the outer edge \cite{kohlbrenner2024gentact}. Sensors can now be distributed within the dermis layer.

$P_{\text{SD}}$ distributes points in a randomized fashion on the dermis surface while respecting user-specified densities by using the Poisson disk sampling algorithm \cite{bridson2007fast}. Cylindrical nodules are instantiated at each point, then shaved to match the dermal contour, and finally translated inward to be fixed within the dermis. Each nodule's radius is scaled proportionally to its closest neighbor and a radius scaling factor. 
This work distinguishes from previous works by embedding the nodules completely within the dermis to prevent capacitive coupling from direct contact with the nodules \cite{kohlbrenner2024gentact}. Direct coupling has a much stronger effect on capacitive sensor readings than the fringe field interaction of nearby objects. By embedding the nodules, the sensors are sensitively tuned for the proximal capacitive effects of objects without being saturated at contact. Next, embedded wires are generated to connect the nodules to a microcontroller for sensing.

$P_{\text{route}}$ creates a graph of vertices connected by edges that signify the possible paths a wire can take. The routing graph resides in the dermis and is created by distributing copies of the weighted vertices $V_{\text{skin}}$ outwards in layers. The designer can use the $A_{\text{route}}$ parameters to specify the number of layers and the thickness between layers. The designer can also pass in a new weighted set of vertices $H_{\text{route}}$ instead of $H_{\text{skin}}$ to specify regions of the dermis that should not have wires. Next, connection ports are distributed along the aforementioned outer edge spline with a user-specified geometry. Finally, all vertices in the routing graph are connected to their neighbors within a specified radius, and the sensors and connection ports are connected to their closest vertex in the routing graph. Wires can now be generated along the graph.

$P_{\text{wire}}$ iteratively generates wires between the sensors and connection ports while avoiding overlapping geometry. Our work takes a similar approach to other 3D printed routing works by framing the generation as a graph-search path planning problem \cite{palma2024capacitive, savage2014series, he2021modelec}. At each iteration, a lightweight $A^*$ algorithm is guided by a heuristic $h$, as seen in \cref{eq:heuristic}, to find candidate wires between every connection port and sensor. The heuristic $h$ can be weighted to favor wire length $L$ or proximity to previously instantiated wires $K$ for wire path $w$ in the set of all possible wires $\mathcal{W}$. 
\begin{equation}
    h(w,a_{\text{mix}})=a_{\text{mix}}  L(w) + (1-a_{\text{mix}})K(w)
    \label{eq:heuristic}
\end{equation}
The vertices in the routing graph occupied by the wire, as well as any edges that come within a set range of the occupied wire, are then removed from the routing graph to prevent overlapping paths. The wires are finally smoothed and given a circular profile for volume. The resulting skin unit is digitally designed and ready for 3D printing.
\begin{figure}
\vspace{2mm}
    \centering
    \includegraphics[width=\linewidth]{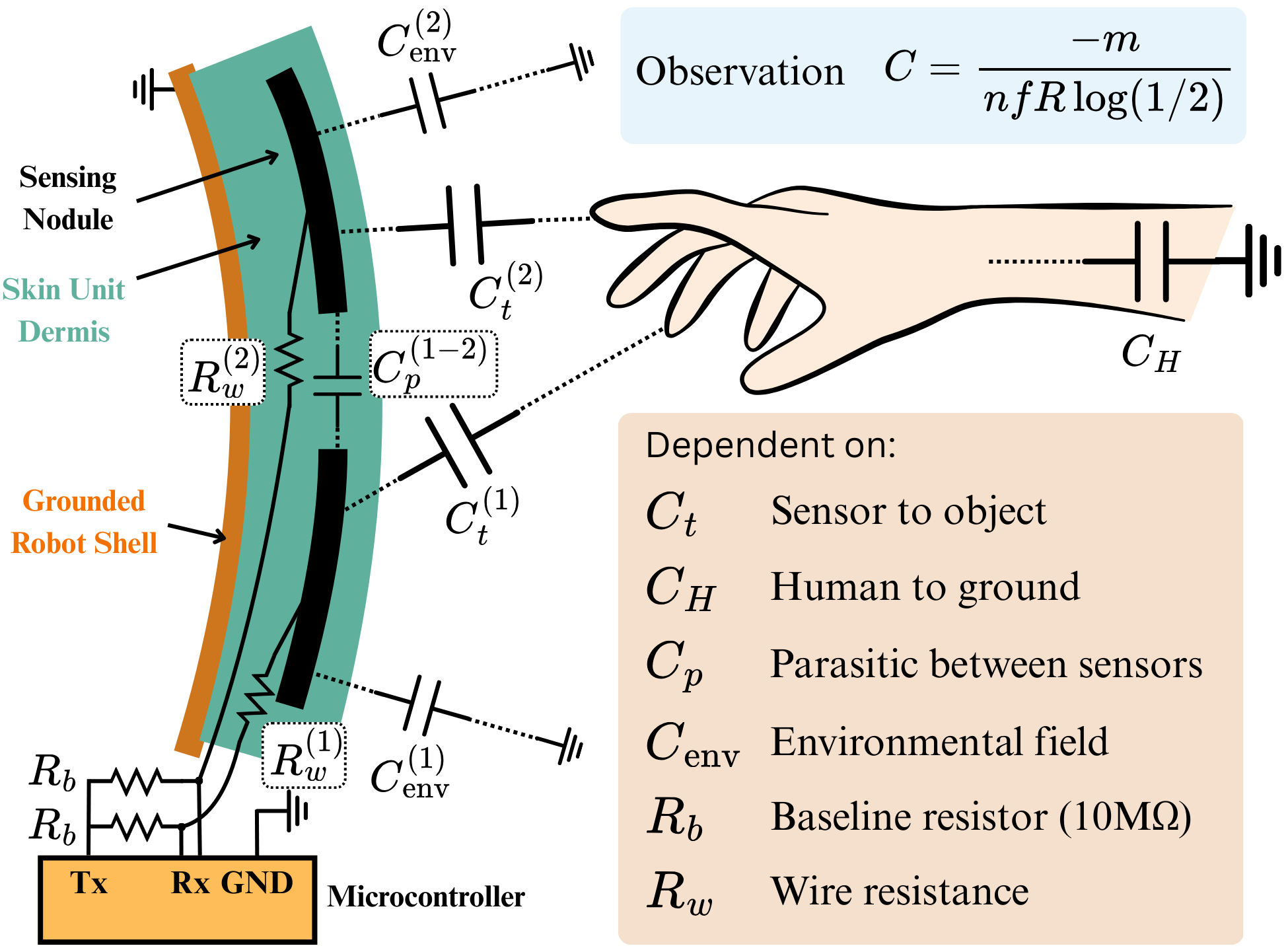}
    \caption{The electrode signals are dependent on the capacitance of the object being detected, parasitic capacitance between sensors, and the environment (which includes the coupling between the sensor and grounded robot shell). The series resistance also impacts sensing quality; however, resistance is assumed constant for a rigid skin.}
    \label{fig:circuit}
    \vspace{-1.5mm}
\end{figure}

\subsection{Fabrication and Calibration}
The skin units are manufactured using multi-material fused deposition modeling (FDM) 3D printing, ensuring a consistent fabrication process for any given skin unit geometry. Polylactic acid (PLA) filament is used to maintain consistent sensor transformations post-fabrication due to its rigidity, and conductive PLA is used to print the wires and capacitive sensors. After printing, a microcontroller can be connected to the sensors via the exposed connection ports.

Calibrating the sensor locations on the robot is streamlined due to the skin units being explicitly designed for the robot's morphology. The flush fit allows them to be snapped on without any adhesives or supports. This snap-on advantage particularly applies to robots with non-developable surface geometry (i.e., geometries that cannot be flattened into a plane without warping), in which the skin units can only be applied in a single configuration on the robot. The kinematic chain of the robot is assumed preserved, where each sensor's pose is defined relative to the frame of the applied link.

\subsection{Self-Capacitance Sensing}
Touch and proximity are measured by the 3D-printed platform via self-capacitance. A capacitive sensor consists of a conductive electrode connected to a measurement circuit that applies an electric potential relative to a reference. This creates an electrostatic field, the distribution of which is perturbed when nearby conductive or dielectric objects approach the electrode. The capacitance is defined as the ratio of stored charge $Q$ to applied potential difference $V$:
\begin{equation}
    C = \frac{Q}{V}
    \label{eq:capacitance-base}
\end{equation}
For the simplified case of an infinite parallel plate capacitor, these two equations simplify to be a function of the transmission volume's permittivity $\epsilon$, the area of overlap between the plates $A$, and the distance between the plates $d$:
\begin{equation}
    C = \frac{\epsilon A}{d}
    \label{eq:cap-plate}
\end{equation}
Parallel plates are an idealized assumption for practical applications of self-capacitance sensors that highlight the sensor's sensitivity to an object's \textsl{distance}, \textsl{shape}, and \textsl{material}.
\begin{figure}
\vspace{1.5mm}
    \centering
    \includegraphics[width=\linewidth]{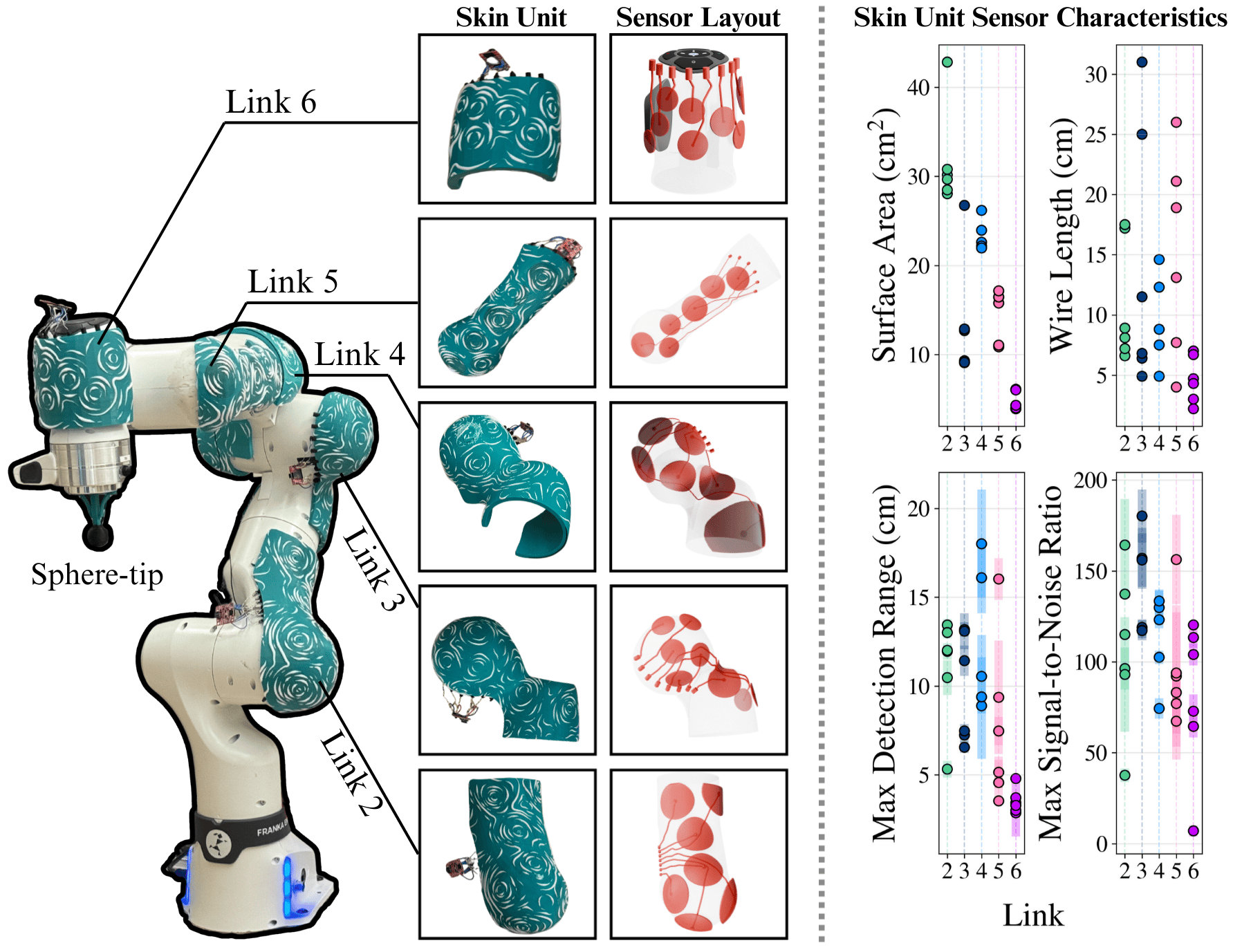}
    \caption{Five unique skin units were printed for the FR3 with the shown properties. It should be noted that the surface areas and wire lengths are a result of the automatic generation process.}
    \label{fig:skins}
    \vspace{-1.5mm}
\end{figure}

The distance between an electrode and an object can be indirectly measured through the time it takes to charge a conductive medium in a resistor-capacitor (RC) circuit. This is achieved in practice by using a microcontroller as both a power supply and a measurement device (\cref{fig:circuit}), where one output pin (Tx) supplies a known voltage $V_\infty$ to an input pin (Rx) that measures the voltage $V$ between the electrode and resistor $R$. The measured voltage approaches the input voltage as the capacitor charges according to \cref{eq:RC}.
\begin{equation}
    V(t) = V_{\infty}(1-e^{-t/RC})
    \label{eq:RC}
\end{equation}
Once $V$ reaches a set threshold, the microcontroller discharges the capacitor. The microcontroller keeps track of how long the charge/discharge process takes by iterating a counter $m$ until the threshold is hit. High-frequency noise can be effectively filtered out of the measurements by averaging an $n$ number of samples, and the resulting capacitance can be calculated using \cref{eq:standard}, where $f$ is the cycle frequency scaled by the CPU clock frequency \cite{Fonseca2023AFP}.
\begin{equation}
    C = \frac{-m}{n f R \log(1/2)}
    \label{eq:standard}
\end{equation}
For a given sensor $i$, as shown in \cref{fig:circuit}, the measured capacitance is a function of the environmental capacitance $C_{\text{env}}^{(i)}$, parasitic capacitance between sensors ($C_p^{(i-j)}$), and the coupling capacitance between nearby objects $C_t^{(i)}$. For our skin units, the resistance of a sensor is the sum of resistances imparted by the generated wire $R_w^{i}$ and a set resistor $R_b$ connecting the Tx and Rx pins of the microcontroller. 

Capacitive sensors can be challenging to work with due to the various environmental impacts on their signal in addition to their parasitic capacitance with other sensors. For applications where coarse precision is sufficient, the parallel plate capacitor assumption can be applied to \cref{eq:standard} to measure object distance $d$ from the electrode online (\cref{eq:distance_meas}). In our study, we add the parameters $k$ and $w$ to empirically tune the sensitivity of the sensors to a given detected object.
\begin{equation}
    d^{w} = \frac{k }{C}= \frac{-k nfR\log(1/2)}{m}
    \label{eq:distance_meas}
\end{equation}

The parameters $k$ and $w$ can be calculated using a linear best fit in logarithmic space using $\log{C=-w\log d + \log{k}}$.

\begin{figure}
\vspace{1.5mm}
    \centering
    \includegraphics[width=\linewidth]{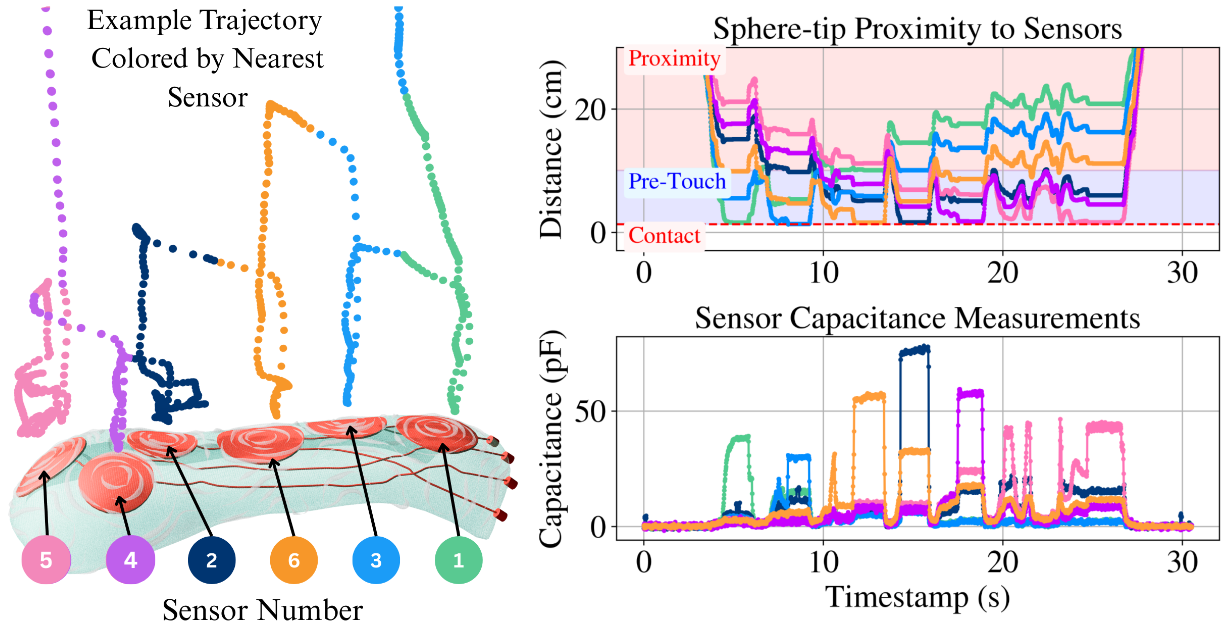}
    \caption{Each collected trajectory included the relative tracking of a conductive sphere and sensed signal for each sensor. The above example trajectory was collected for link five, where the nearest sensor color is assigned to each timestamp.}
    \label{fig:experiment}
    \vspace{-1.5mm}
    
\end{figure}

\begin{figure*}
    \vspace{2mm}
    \centering
    \includegraphics[width=\linewidth]{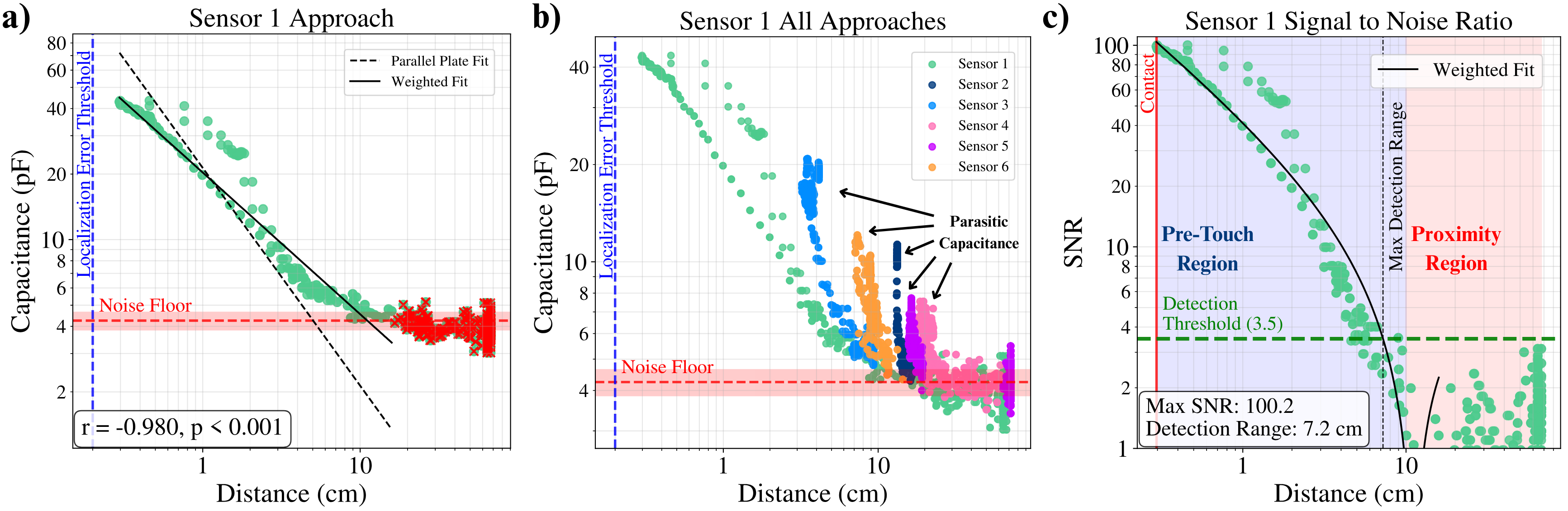}
    \caption{Each chart is an example of a single sensor's characterization for one trajectory in the experiment. a) The parallel plate assumption is fit to the isolated capacitance values for the sensor's respective approach. Samples collected at a further distance than the closest sample under the noise floor were assumed to be noise and discarded from fitting. b) The parasitic influences of the nearby sensors are shown for a full trajectory where all sensors were approached. c) The detection range was determined as the distance the weighted fit line intercepts the SNR detection threshold of 3.5.}
    \label{fig:characterization}
    \vspace{-1.5mm}
\end{figure*}

\begin{figure}
    \centering
    \includegraphics[width=\linewidth]{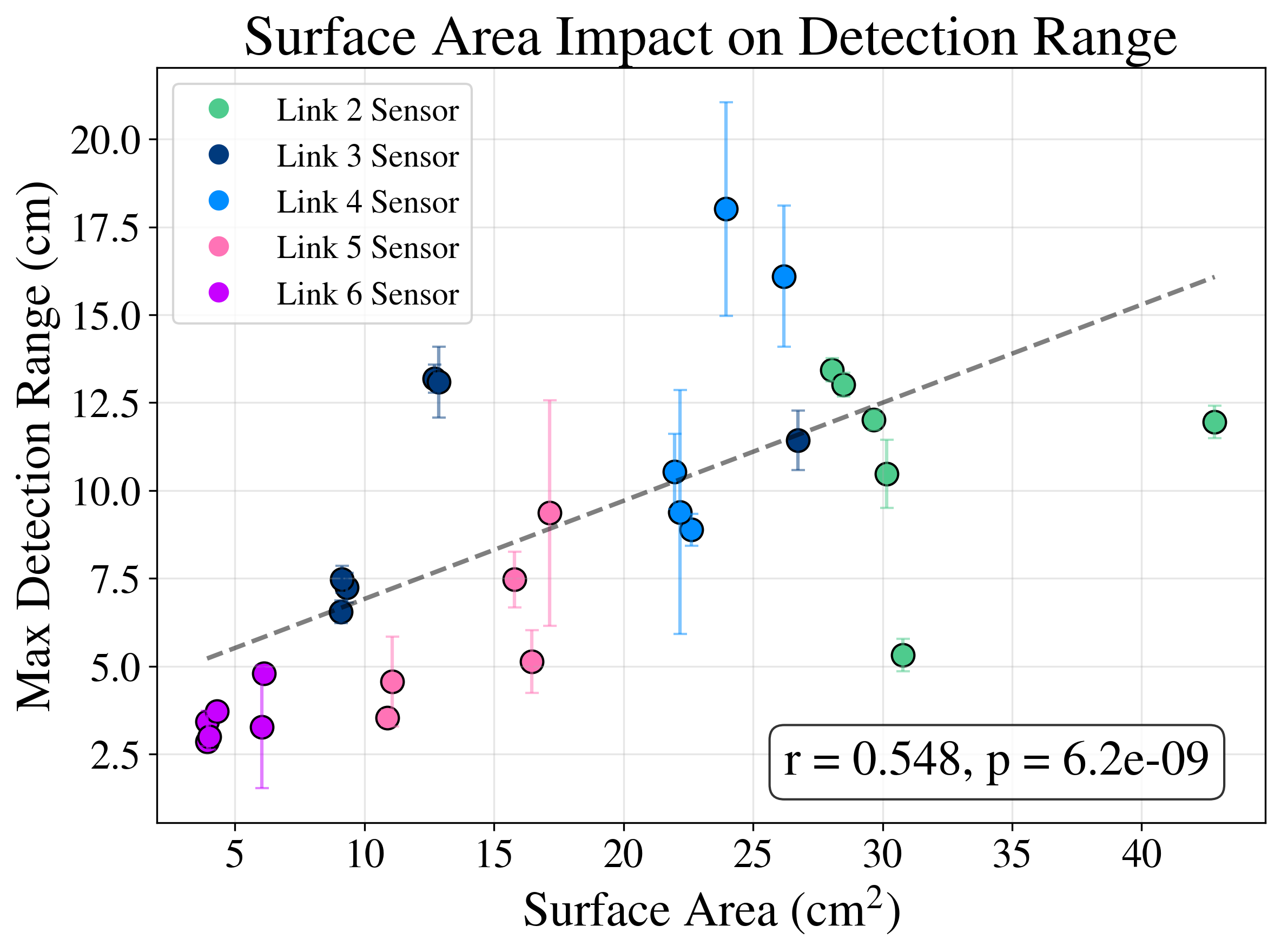}
    \caption{The maximum detection ranges for each sensor in each link are shown in ascending order by surface area. We recorded that surface area had a moderate impact on the maximum detection range when considering all 17 collected trajectories. }
    \label{fig:snr_v_area}
    \vspace{-1.5mm}
\end{figure}

\subsection{Perisensory Space Mapping}
To characterize the space around the robot, we construct a PSS map that links capacitance measurements to object positions. This map describes regions of space where the skin can deliver reliable position estimates for a consistent object shape and material, enabling contact anticipation. 

We train an ensemble of feed-forward multi-layer perceptrons (MLPs) on labeled pairs of capacitance readings and object positions $p = [x,y,z]^T$. Each MLP is trained on a randomly shuffled bootstrap sample of the data to encourage diversity \cite{Lakshminarayanan2016SimpleAS}. For a given input, the ensemble produces a distribution of predicted positions. We assume this distribution is Gaussian and represent it by a mean $\mu_p$ and standard deviation $\sigma_p$, computed relative to the skin unit’s origin point \cite{ganaie2022ensemble}. The skin unit origin points used in this work coincide with the origin point of the applied link in the robot’s kinematic chain, whereas the individual sensor origin points are defined from the nodule centers. To prevent overconfidence, the predicted uncertainty is scaled using a regression fit between $\sigma_p$ and the mean prediction error $e_p = ||\mu_p - p||_2$ on a held-out validation set.

We then project these predictions onto a grid surrounding the robot. Random capacitance values are sampled and passed through the ensemble, producing position estimates with uncertainties. Each prediction is assigned to the nearest grid point, and each grid point stores the mean uncertainty of the predictions assigned to it. With sufficient sampling, the resulting grid visualizes the effective proximity-sensitive regions of the skin, defining the robot’s PSS.

\section{Results}

We analytically characterized a set of unique skin units, mapped out their effective sensing regions, and deployed the skin units in a collision avoidance task to verify that the space surrounding a robot can be used for contact anticipation with our 3D printed artificial skins.

\begin{figure*}
    \vspace{2mm}
    \centering
    \includegraphics[width=\linewidth]{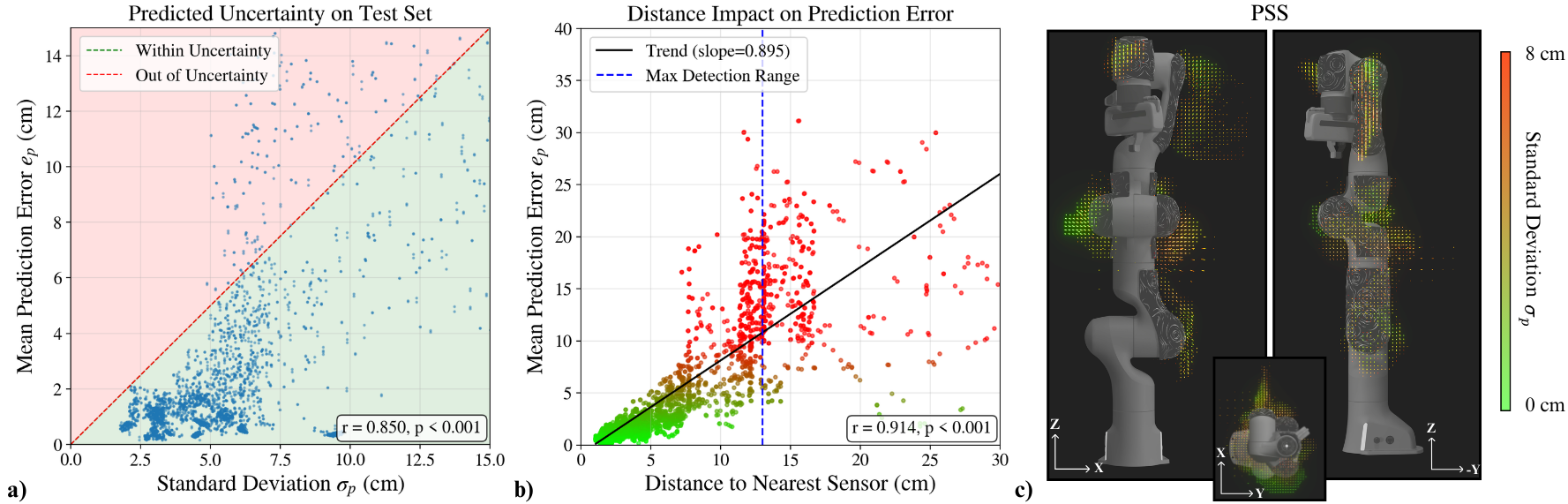}
    \caption{An example test set is demonstrated in a and b for a single skin unit (Link 3). a) The ensemble's prediction of uncertainty was strongly correlated to the true error and reliable for small values of $\sigma_p$. b) The model's prediction accuracy was strongly correlated to distance from the sensors until the test object was out of the sensing range. c) The ensemble's uncertainty predictions revealed the robot's effective PSS.}
    \label{fig:predictions}
    \vspace{-1.5mm}
\end{figure*}

\subsection{Experimental Setup and Data Collection}

We designed, fabricated, and evaluated five unique skin units for the Franka Research 3 robot arm as seen in \cref{fig:skins}. Each skin unit was printed on a Prusa XL using Protopasta Electrically Conductive PLA for the conductive elements. Small cylindrical connection ports were connected to the wires of an ESP32-C6 microcontroller through heat treatment. The Sparkfun ESP32-C6 Qwiic-Dev board used in this study ran at a clock frequency of 160 MHz and communicated data via serial bus at 15-30 Hz when unsaturated. 

Our characterization testing experiment consisted of hovering a grounded $\qty{25}{mm}$ diameter conductive PLA sphere over each of the five skin units. The sphere-tip was attached to the end effector of the FR3 robot arm (as seen in \cref{fig:skins}) to track its position in Cartesian space over time using forward kinematics. Each of the skin units was secured on a table directly in front of the FR3 and manually localized relative to the robot arm using hand measurements. Three distributed distal cameras captured the experiments and were used to check for consistency with all manual calibrations. We conservatively estimate that a $\pm \qty{2}{mm}$ localization error was induced during the hand calibration process. The robot arm was teleoperated for each experiment to hover directly over each sensor before touching their centers; however, there was not necessarily a direct straight-line trajectory connecting the hover position to the contact position. For characterizing the sensors, we classified each data point in a given trajectory to the individual sensor that was being approached at a given time, then isolated the signal for each sensor's approach. \cref{fig:experiment} shows an example trajectory collected for link five. For mapping the PSS, we passed the full set of capacitance-position pairs directly to the ensemble. Three to six separate experiments conducted at room temperature were recorded for each skin unit.

\subsection{Sensor Characterization}
\label{sec:characterization}

We demonstrate that our 3D printed skins are sufficient for proximity detection through an analytical assessment of sensitivity, signal-to-noise ratio, and detection range for each skin unit. 
We found that our experimental data were well described by the following parallel plate assumption (\cref{eq:distance_meas}) with sensitivity $w$ and constant $k$ when the detected object does not change shape or permittivity.
\cref{fig:characterization}-a shows the isolated approaches plotted in logarithmic space, along with the expected sensitivity of a perfect parallel plate capacitor ($w=1$) for comparison. We observed that each sensor was strongly inversely correlated (using Pearson's correlation) with distance and was well represented by a linear sensitivity of $0.4\leq w<1$. Parasitic coupling between sensors was observed in the spikes captured by individual sensors while adjacent sensors were being approached, as seen in \cref{fig:characterization}-b. Filtering out the parasitic capacitance was not found to be necessary for mapping the PSS in \cref{sec:pps_results}. Next, we quantified the maximum detection range and signal-to-noise ratio (SNR) for each sensor.

SNR is a metric used for capacitive sensors to evaluate the strength of a sensor's active signal due to the presence of an object against its inactive signal due to background noise. The following formula was used to calculate SNR \cite{2010AN1T}:
\begin{equation}
    \label{eq:snr}
    SNR =\frac{\left|\mu_n-\mu\right|}{\sigma_n}
\end{equation}
A two-second calibration period occurred at the beginning and end of each experiment to take the mean inactive signal $\mu_n$ and standard deviation $\sigma_n$. The maximum detection range is defined in this work as the distance where the curve fit, after being converted to SNR using \cref{eq:snr}, crosses the minimum SNR threshold of 3.5 for reliable detection \cite{2010AN1T}. \cref{fig:characterization}-c shows an example of the SNR levels for one sensor for each measured distance from the sphere-tip object. The maximum SNR during contact and maximum detection ranges are reported for each sensor in \cref{fig:skins}. The large variability in maximum detection ranges---approximately 2.6 to 18 cm---was found to be in part caused by design factors such as sensor surface area (\cref{fig:snr_v_area}). Next, we used the characterized skin units to map out the PSS for the FR3.

\subsection{Perisensory Space Mapping}
\label{sec:pps_results}

We mapped the PSS for the FR3 using an ensemble of MLPs and assessed its accuracy at localizing the sphere-tip object as it followed trajectories unseen in its training data. The ensemble hyperparameters were manually tuned to maximize training results. An ensemble of 100 MLPs was trained for each skin unit with two dropout hidden layers of size 64 and was trained using the ADAM optimizer \cite{kingma2014adam} for 40 epochs at a learning rate of $10^{-4}$. Each model in the ensemble was trained on a random subset of 50\% of the training data. The first two seconds of each trajectory, before the sphere-tip approached the skin units, were used as the baseline capacitance values and subtracted from the rest of the training data. We then compared $\sigma_p$ to $e_p$ from test trajectories (Example in \cref{fig:predictions}-a) to verify that the ensemble predictions are applicable to nearby regions unseen in the training data. 

We observed a strong correlation between $e_p$ and $\sigma_p$ (\cref{fig:predictions}-a), with a tendency for the models to be underconfident when predicting small errors (approx. $\leq \qty{8}{cm}$) and overconfident when predicting errors larger than their respective maximum detection ranges. A dramatic drop off in prediction accuracy can also be observed in \cref{fig:predictions}-b when the target is further than the maximum detection range. These results signify that the learned correlation between space and localization uncertainties is representative of the true prediction accuracy, and that the model is applicable to nearby regions unseen in the training data. Thus, we can filter predictions by their uncertainties to retrieve the usable space for contact anticipation.

We sampled 50,000 random combinations of raw capacitance values at ranges found within the training data to collect a diverse set of points to use in the PSS mapping. \cref{fig:predictions}-c presents the PSS map as grid points separated by $\qty{1}{cm}$ and colored by $\sigma_p$. Between 0-\qty{8}{cm} was found to be a sufficient cutoff uncertainty for filtering out samples outside of the usable space. Regions close to the robot skin units generally had lower predicted uncertainties than regions further from the robot. It was also observed that regions between sensors had lower uncertainties, which concur with expectations that objects within the detection range of multiple sensors should be localized with higher accuracy.

\subsection{Collision Avoidance Control}
We adapted a Cartesian velocity controller for the FR3 to interface with our skin units, demonstrating the use of proximity sensing for online collision avoidance \cite{escobedo2021contact}. The PSS was used to inform where obstacles can be detected relative to the robot. The avoidance controller calculates repulsive vectors that slow down and push away the end-effector from detected obstacles. \cref{eq:distance_meas} was used to estimate obstacle proximity and was projected along the normal vector of each sensor. We found that adding a slowly adapting PID controller that zeros out lingering capacitance values eliminated drift without noticeable detriment to its sensing ability, allowing for prolonged deployment.

\begin{figure}
    \vspace{2mm}
    \centering
    \includegraphics[width=\linewidth]{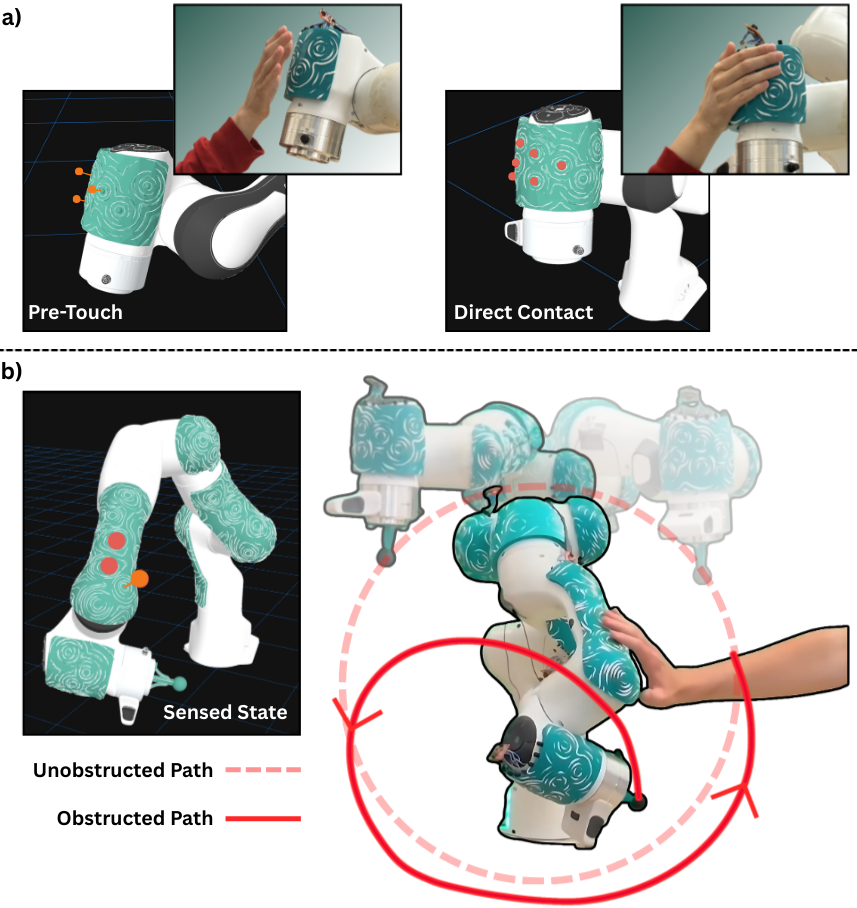}
    \caption{a) Examples of pre-touch and direct contact detection are represented as a point cloud using equation \cref{eq:distance_meas}. b) The FR3 is tasked to trace out a circle in free space while the detection of a human hand informs a collision-compliant controller to both avoid the obstacle and reduce its velocity near it. }
    \label{fig:collision-avoidance}
    \vspace{-1.5mm}
\end{figure}

\cref{fig:collision-avoidance}-a shows an example of the controller detecting pre-touch proximity and direct contact stimuli over the embedded sensors. The positions of each estimate relative to the base of the robot were extracted using forward kinematics, then passed into the avoidance controller's repulsive vector calculation to avoid the obstructions. \cref{fig:collision-avoidance}-b plots the trajectory for two cycles of the FR3 when a human hand was inserted in the path of the robot as it was tracing out a circle. The controller successfully reduced its velocity and avoided the obstructed region of the path before resuming its unobstructed cycle. An example of the collision avoidance controller reacting to obstacles are available in the supporting video attachment.

\section{Conclusion}
In this work, we introduce \textit{GenTact-Prox}, a scalable, low-cost ($<\$25$, USD), and fully 3D-printed artificial skin that integrates tactile and proximity sensing over a robot’s body. The skin resolution is customizable in detection range for sensing nearby objects. The artificial skin was specifically used for contact anticipation in this study, but can also be extended to a broad range of proximity sensing applications. To characterize the sensing capabilities of our sensors, we present a data-driven framework to map the Perisensory Space, which is the body-centric volume of space around a robot in which the sensors can detect nearby objects. Finally, our evaluation of an FR3 robot fully covered in \textit{GenTact-Prox} skin units was directly leveraged for online collision-compliant control.

Limitations of this work include unevaluated impacts due to a changing environment or nearby object shape. In practice, we found that a tuned parallel plate assumption \cref{eq:distance_meas} was sufficient for coarse proximity detection, and we anticipate that higher accuracy localization can be achieved with additional perceptual software that estimates nearby object shape and material.

Looking ahead, several steps remain for advancing toward an artificial PPS in robots. First, broader data collection across multiple robots and skin configurations is needed to evaluate how well PSS models generalize beyond a single deployment. Second, the PPS is known to adapt in response to tool use and physiological changes which was unexplored in this work \cite{holmes2004body}. Third, integrating additional sensing modalities such as vision or force feedback may enrich PSS mappings and improve robustness in unstructured environments. Progress in these directions is essential for moving from static, sensor-level receptive fields toward a dynamic, body-centric PPS that supports anticipatory and safe interaction in general unstructured environments.

\bibliographystyle{IEEEtran}
\bibliography{references}

@article{cheng2019comprehensive,
  title={A comprehensive realization of robot skin: Sensors, sensing, control, and applications},
  author={Cheng, Gordon and Dean-Leon, Emmanuel and Bergner, Florian and Olvera, Julio Rogelio Guadarrama and Leboutet, Quentin and Mittendorfer, Philipp},
  journal={Proceedings of the IEEE},
  volume={107},
  number={10},
  pages={2034--2051},
  year={2019},
  publisher={IEEE}
}

@article{zhou2023tacsuit,
  title={TacSuit: A wearable large-area, bioinspired multimodal tactile skin for collaborative robots},
  author={Zhou, Yanmin and Zhao, Jiangang and Lu, Ping and Wang, Zhipeng and He, Bin},
  journal={IEEE Transactions on Industrial Electronics},
  volume={71},
  number={2},
  pages={1708--1717},
  year={2023},
  publisher={IEEE}
}

@inproceedings{kohlbrenner2024gentact,
  title={GenTact Toolbox: A Computational Design Pipeline to Procedurally Generate Context-Driven 3D Printed Whole-Body Artificial Skins},
  author={Kohlbrenner, Carson and Escobedo, Caleb and Bae, S Sandra and Dickhans, Alexander and Roncone, Alessandro},
  booktitle={2025 IEEE International Conference on Robotics and Automation (ICRA)},
  pages={4716--4722},
  year={2025},
  organization={IEEE}
}

@article{yang2024digital,
  title={A Digital Twin-Based Large-Area Robot Skin System for Safer Human-Centered Healthcare Robots Toward Healthcare 4.0},
  author={Yang, Geng and Ye, Zhiqiu and Wu, Haiteng and Li, Chen and Wang, Ruohan and Kong, Depeng and Hou, Zeyang and Wang, Huafen and Huang, Xiaoyan and Pang, Zhibo and others},
  journal={IEEE Transactions on Medical Robotics and Bionics},
  volume={6},
  number={3},
  pages={1104--1115},
  year={2024},
  publisher={IEEE}
}

@inproceedings{fan2021aurasense,
  title={Aurasense: Robot collision avoidance by full surface proximity detection},
  author={Fan, Xiaoran and Simmons-Edler, Riley and Lee, Daewon and Jackel, Larry and Howard, Richard and Lee, Daniel},
  booktitle={2021 IEEE/RSJ International Conference on Intelligent Robots and Systems (IROS)},
  pages={1763--1770},
  year={2021},
  organization={IEEE}
}

@inproceedings{escobedo2021contact,
  title={Contact anticipation for physical human--robot interaction with robotic manipulators using onboard proximity sensors},
  year={2021},
  author={Escobedo, Caleb and Strong, Matthew and West, Mary and Aramburu, Ander and Roncone, Alessandro},
  booktitle={IEEE IROS},
}

@article{corniani2020tactile,
  title={Tactile innervation densities across the whole body},
  author={Corniani, Giulia and Saal, Hannes P},
  journal={Journal of Neurophysiology},
  volume={124},
  number={4},
  pages={1229--1240},
  year={2020},
  publisher={American Physiological Society Bethesda, MD}
}

@inproceedings{watanabe2021self,
  title={Self-contained kinematic calibration of a novel whole-body artificial skin for human-robot collaboration},
  author={Watanabe, Kandai and Strong, Matthew and West, Mary and Escobedo, Caleb and Aramburu, Ander and Kodur, Krishna Chaitanya and Roncone, Alessandro},
  booktitle={2021 IEEE/RSJ International Conference on Intelligent Robots and Systems (IROS)},
  pages={1778--1785},
  year={2021},
  organization={IEEE}
}

@article{palma2024capacitive,
  title={Capacitive Touch Sensing on General 3D Surfaces},
  author={Palma, Gianpaolo and Pourjafarian, Narges and Steimle, J{\"u}rgen and Cignoni, Paolo},
  journal={ACM Transactions on Graphics},
  volume={43},
  number={4},
  pages={1--20},
  year={2024},
  publisher={ACM New York, NY, USA}
}

@article{serino2022peripersonal,
  title={Peripersonal space (PPS)},
  author={Serino, Andrea},
  journal={The Routledge Handbook of Bodily Awareness},
  pages={171--184},
  year={2022},
  publisher={Routledge}
}

@article{bae2023computational,
  title={A Computational Design Pipeline to Fabricate Sensing Network Physicalizations},
  author={Bae, S Sandra and Fujiwara, Takanori and Ynnerman, Anders and Do, Ellen Yi-Luen and Rivera, Michael L and Szafir, Danielle Albers},
  journal={IEEE Transactions on Visualization and Computer Graphics},
  year={2023},
  publisher={IEEE}
}

@article{bridson2007fast,
  title={Fast Poisson disk sampling in arbitrary dimensions.},
  author={Bridson, Robert},
  journal={SIGGRAPH sketches},
  volume={10},
  number={1},
  pages={1},
  year={2007}
}

@inproceedings{roncone2015learning,
  title={Learning peripersonal space representation through artificial skin for avoidance and reaching with whole body surface},
  author={Roncone, Alessandro and Hoffmann, Matej and Pattacini, Ugo and Metta, Giorgio},
  booktitle={2015 IEEE/RSJ International Conference on Intelligent Robots and Systems (IROS)},
  pages={3366--3373},
  year={2015},
  organization={IEEE}
}

@article{roncone2016peripersonal,
  title={Peripersonal space and margin of safety around the body: learning visuo-tactile associations in a humanoid robot with artificial skin},
  author={Roncone, Alessandro and Hoffmann, Matej and Pattacini, Ugo and Fadiga, Luciano and Metta, Giorgio},
  journal={PloS one},
  volume={11},
  number={10},
  pages={e0163713},
  year={2016},
  publisher={Public Library of Science San Francisco, CA USA}
}

@inproceedings{rupavatharam2023ambisense,
  title={AmbiSense: Acoustic Field Based Blindspot-Free Proximity Detection and Bearing Estimation},
  author={Rupavatharam, Siddharth and Fan, Xiaoran and Escobedo, Caleb and Lee, Daewon and Jackel, Larry and Howard, Richard and Prepscius, Colin and Lee, Daniel and Isler, Volkan},
  booktitle={2023 IEEE/RSJ International Conference on Intelligent Robots and Systems (IROS)},
  pages={5974--5981},
  year={2023},
  organization={IEEE}
}

@article{Fonseca2023AFP,
  title={A Flexible Piezoresistive/Self-Capacitive Hybrid Force and Proximity Sensor to Interface Collaborative Robots},
  author={Diogo Fonseca and Mohammad Safeea and Pedro Neto},
  journal={IEEE Transactions on Industrial Informatics},
  year={2023},
  volume={19},
  pages={2485-2495},
  url={https://api.semanticscholar.org/CorpusID:248760168}
}

@inproceedings{Lakshminarayanan2016SimpleAS,
  title={Simple and Scalable Predictive Uncertainty Estimation using Deep Ensembles},
  author={Balaji Lakshminarayanan and Alexander Pritzel and Charles Blundell},
  booktitle={Neural Information Processing Systems},
  year={2016},
  url={https://api.semanticscholar.org/CorpusID:6294674}
}

@misc{2010AN1T,
  title={AN 1334 Techniques for Robust Touch Sensing Design},
  author={},
  year={2010},
  url={https://api.semanticscholar.org/CorpusID:11888772}
}

@article{giovinazzo2024cyskin,
  title={From CySkin to ProxySKIN: Design, implementation and testing of a multi-modal robotic skin for human--robot interaction},
  author={Giovinazzo, Francesco and Grella, Francesco and Sartore, Marco and Adami, Manuela and Galletti, Riccardo and Cannata, Giorgio},
  journal={Sensors},
  volume={24},
  number={4},
  pages={1334},
  year={2024},
  publisher={MDPI}
}

@inproceedings{savage2014series,
  title={A series of tubes: adding interactivity to 3D prints using internal pipes},
  author={Savage, Valkyrie and Schmidt, Ryan and Grossman, Tovi and Fitzmaurice, George and Hartmann, Bj{\"o}rn},
  booktitle={Proceedings of the 27th annual ACM symposium on User interface software and technology},
  pages={3--12},
  year={2014}
}

@inproceedings{schmitz2015capricate,
  title={Capricate: A fabrication pipeline to design and 3D print capacitive touch sensors for interactive objects},
  author={Schmitz, Martin and Khalilbeigi, Mohammadreza and Balwierz, Matthias and Lissermann, Roman and M{\"u}hlh{\"a}user, Max and Steimle, J{\"u}rgen},
  booktitle={Proceedings of the 28th Annual ACM Symposium on User Interface Software \& Technology},
  pages={253--258},
  year={2015}
}

@article{cooke2003complex,
  title={Complex movements evoked by microstimulation of the ventral intraparietal area},
  author={Cooke, Dylan F and Taylor, Charlotte SR and Moore, Tirin and Graziano, Michael SA},
  journal={Proceedings of the National Academy of Sciences},
  volume={100},
  number={10},
  pages={6163--6168},
  year={2003},
  publisher={The National Academy of Sciences}
}

@article{ackerley2014touch,
  title={Touch perceptions across skin sites: differences between sensitivity, direction discrimination and pleasantness},
  author={Ackerley, Rochelle and Carlsson, Ida and Wester, Henric and Olausson, H{\aa}kan and Backlund Wasling, Helena},
  journal={Frontiers in behavioral neuroscience},
  volume={8},
  pages={54},
  year={2014},
  publisher={Frontiers Media SA}
}

@article{holmes2004body,
  title={The body schema and multisensory representation (s) of peripersonal space},
  author={Holmes, Nicholas P and Spence, Charles},
  journal={Cognitive processing},
  volume={5},
  number={2},
  pages={94--105},
  year={2004},
  publisher={Springer}
}

@article{he2021modelec,
  title={ModElec: A design tool for prototyping physical computing devices using conductive 3D printing},
  author={He, Liang and Wittkopf, Jarrid A and Jun, Ji Won and Erickson, Kris and Ballagas, Rafael Tico},
  journal={Proceedings of the ACM on Interactive, Mobile, Wearable and Ubiquitous Technologies},
  volume={5},
  number={4},
  pages={1--20},
  year={2021},
  publisher={ACM New York, NY, USA}
}

@article{kim2024armor,
  title={Armor: Egocentric perception for humanoid robot collision avoidance and motion planning},
  author={Kim, Daehwa and Srouji, Mario and Chen, Chen and Zhang, Jian},
  journal={arXiv preprint arXiv:2412.00396},
  year={2024}
}

@inproceedings{burstyn2015printput,
  title={Printput: Resistive and capacitive input widgets for interactive 3D prints},
  author={Burstyn, Jesse and Fellion, Nicholas and Strohmeier, Paul and Vertegaal, Roel},
  booktitle={IFIP Conference on Human-Computer Interaction},
  pages={332--339},
  year={2015},
  organization={Springer}
}

@article{Ye2020ARO,
  title={A Review on Applications of Capacitive Displacement Sensing for Capacitive Proximity Sensor},
  author={Yong Ye and Chiya Zhang and Chunlong He and Xi Vincent Wang and Jianjun Huang and Jiahao Deng},
  journal={IEEE Access},
  year={2020},
  volume={8},
  pages={45325-45342},
  url={https://api.semanticscholar.org/CorpusID:212705478}
}

@article{bertoni2025computational,
  title={Computational Models of Peripersonal Space Representation},
  author={Bertoni, Tommaso and Chauhan, Ishan-Singh J and Noel, Jean-Paul and Serino, Andrea},
  journal={Physics of Life Reviews},
  year={2025},
  publisher={Elsevier}
}

@article{rozlivek2024harmonious,
  title={HARMONIOUS--Human-like reactive motion control and multimodal perception for humanoid robots},
  author={Rozlivek, Jakub and Roncone, Alessandro and Pattacini, Ugo and Hoffmann, Matej},
  journal={IEEE Transactions on Robotics},
  year={2024},
  publisher={IEEE}
}

@inproceedings{schmitz2019trilaterate,
  title={./trilaterate: A fabrication pipeline to design and 3D print hover-, touch-, and force-sensitive objects},
  author={Schmitz, Martin and Stitz, Martin and M{\"u}ller, Florian and Funk, Markus and M{\"u}hlh{\"a}user, Max},
  booktitle={Proceedings of the 2019 CHI conference on human factors in computing systems},
  pages={1--13},
  year={2019}
}

@article{xiao2025fully,
  title={Fully 3D-Printed Soft Capacitive Sensor of High Toughness and Large Measurement Range},
  author={Xiao, Fei and Wei, Zhuoheng and Xu, Zhipeng and Wang, Hao and Li, Jisen and Zhu, Jian},
  journal={Advanced Science},
  volume={12},
  number={8},
  pages={2410284},
  year={2025},
  publisher={Wiley Online Library}
}

@inproceedings{taylor2024fully,
  title={Fully 3D printable Robot Hand and Soft Tactile Sensor based on Air-pressure and Capacitive Proximity Sensing},
  author={Taylor, Sean and Park, Kyungseo and Yamsani, Sankalp and Kim, Joohyung},
  booktitle={2024 IEEE International Conference on Robotics and Automation (ICRA)},
  pages={18100--18105},
  year={2024},
  organization={IEEE}
}

@article{ganaie2022ensemble,
  title={Ensemble deep learning: A review},
  author={Ganaie, Mudasir A and Hu, Minghui and Malik, Ashwani Kumar and Tanveer, Muhammad and Suganthan, Ponnuthurai N},
  journal={Engineering Applications of Artificial Intelligence},
  volume={115},
  pages={105151},
  year={2022},
  publisher={Elsevier}
}

@article{kingma2014adam,
  title={Adam: A method for stochastic optimization},
  author={Kingma, Diederik P and Ba, Jimmy},
  journal={arXiv preprint arXiv:1412.6980},
  year={2014}
}

@misc{poignant2024teleoperationroboticmanipulatorperipersonal,
      title={Teleoperation of a robotic manipulator in peri-personal space: a virtual wand approach}, 
      author={Alexis Poignant and Guillaume Morel and Nathanaël Jarrassé},
      year={2024},
      eprint={2406.09309},
      archivePrefix={arXiv},
      primaryClass={cs.RO},
      url={https://arxiv.org/abs/2406.09309}, 
}
\end{document}